\documentclass[10pt,twocolumn,letterpaper]{article}

\usepackage[pagenumbers]{cvpr} 
\usepackage{bm}
\usepackage{microtype}
\usepackage{graphicx}
\usepackage{booktabs} 

\usepackage[dvipsnames]{xcolor}
\usepackage{algorithm}
\usepackage{algorithmic}

\usepackage{amsmath}
\usepackage{amssymb}
\usepackage{mathtools}
\usepackage{amsthm}
\usepackage{multirow} 
\usepackage{bm}

\definecolor{cvprblue}{rgb}{0.21,0.49,0.74}
\usepackage[pagebackref,breaklinks,colorlinks,allcolors=cvprblue]{hyperref}

\def\paperID{*****} 
\def\confName{CVPR}
\def\confYear{2025}

\title{DWTNeRF: Boosting Few-shot Neural Radiance Fields via Discrete Wavelet Transform}

\author{Hung Nguyen \qquad Runfa Blark Li \qquad Truong Nguyen\\
Video Processing Lab, UC San Diego\\
{\tt\small \{hun004, runfa, tqn001\}@ucsd.edu}
}

\begin{document}
\maketitle
\begin{abstract}
Neural Radiance Fields (NeRF) has achieved superior performance in novel view synthesis and 3D scene representation, but its practical applications are hindered by slow convergence and reliance on dense training views. To this end, we present DWTNeRF, a unified framework based on Instant-NGP's fast-training hash encoding. It is coupled with regularization terms designed for few-shot NeRF, which operates on sparse training views. Our DWTNeRF additionally includes a novel Discrete Wavelet loss that allows explicit prioritization of low frequencies directly in the training objective, reducing few-shot NeRF's overfitting on high frequencies in earlier training stages. We also introduce a model-based approach, based on multi-head attention, that is compatible with INGP, which are sensitive to architectural changes. On the 3-shot LLFF benchmark, DWTNeRF outperforms Vanilla INGP by 15.07\% in PSNR, 24.45\% in SSIM and 36.30\% in LPIPS. Our approach encourages a re-thinking of current few-shot approaches for fast-converging implicit representations like INGP or 3DGS.
\end{abstract}    
\section{Introduction}
\label{sec:intro}

\begin{figure}
\vskip 0.2in
\begin{center}
\centerline{\includegraphics[width=\columnwidth]{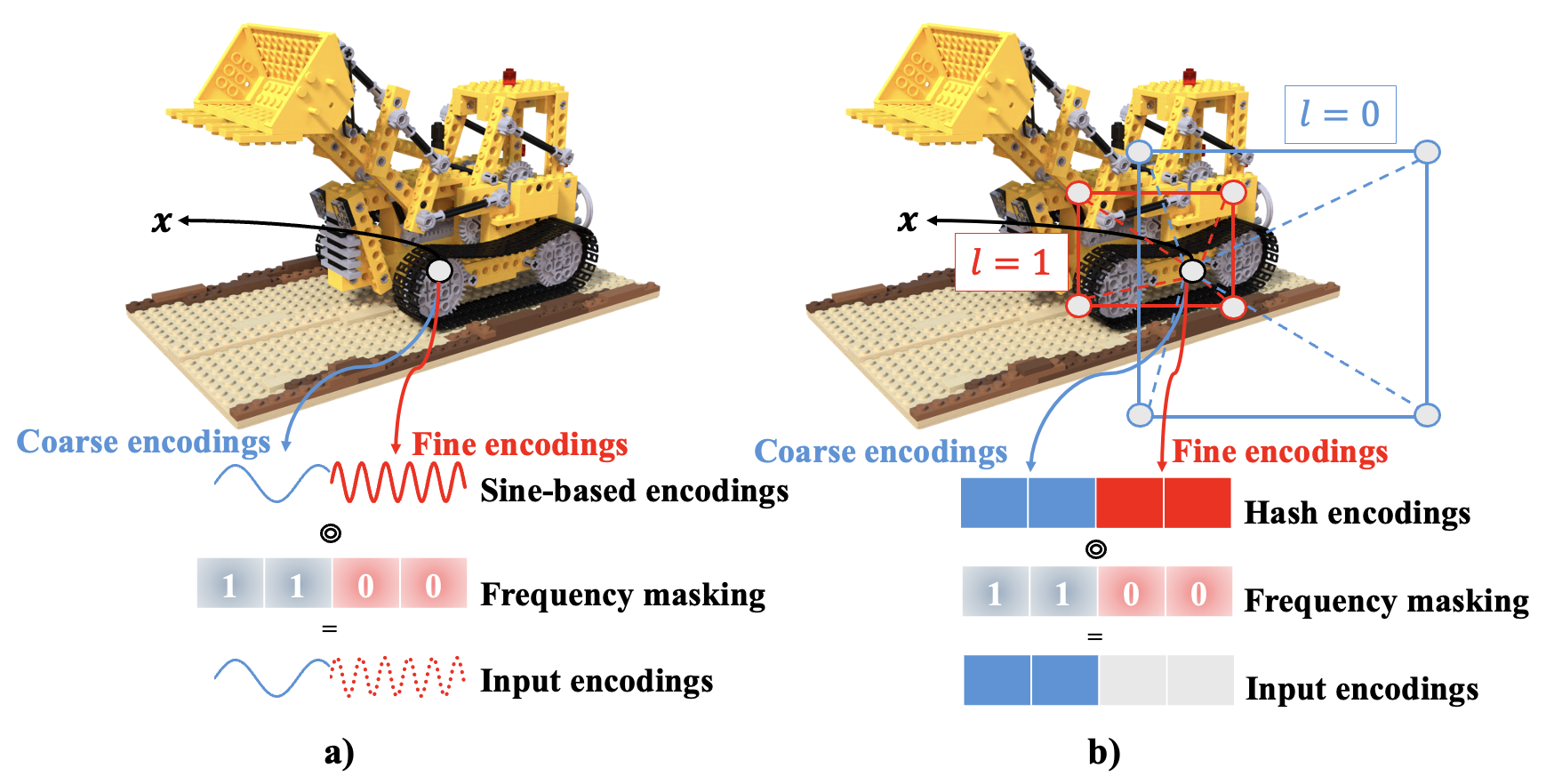}}
\caption{Comparison of frequency masking as applied to sine-based encoding (a) and multi-resolution hash encoding (b). To be compatible with frequency masking \cite{FreeNeRF}, the later portions of the encoded inputs $\gamma(\mathbf{x})$ must strictly correspond to high frequencies. For the hash encoding, those input portions correspond to fine resolutions, which are not interchangeable with high frequencies.}
\label{fig_freq_diff}
\end{center}
\vskip -0.2in
\end{figure}

\quad NeRF \cite{NeRF} have emerged as a powerful method for reconstructing 3D scenes from a set of 2D images. This can generate photorealistic views from novel angles. Its properties have enabled applications in medical imaging \cite{MedNeRF, CuNeRF}, robotics \cite{NeRF-robotics, NeRF-robotics2}, virtual reality \cite{NeRF-VR, HumaNeRF}, autonomous driving \cite{NeRF-driving, NeuRAD}, and many more.

However, a disadvantage of NeRF is its slow convergence, which can take several hours to render a scene \cite{plenoxels}. Besides, it requires a dense set of training views \cite{FlipNeRF}. Those problems can hinder NeRF's practical uses. For the former problem, INGP \cite{INGP} introduced a multi-resolution hash encoding that dramatically reduces training time. However, to the best of our knowledge, limited research has been conducted on few-shot INGP. Therefore, our work will focus on addressing this gap.

INGP has distinct characteristics that make some recent few-shot approaches incompatible. Firstly, its positional encoding is based on multi-resolution. This differs from the sine-based encoding \cite{spectral} in Vanilla NeRF and similar works, which \textit{explicitly} map inputs to high frequencies for learning intricate details. The sine-based encoding supports frequency masking \cite{FreeNeRF}, a regularization strategy that element-wise masks high-frequency inputs early in training to prevent overfitting. However, applying this to INGP, as in Figure \ref{fig_freq_diff}, assumes fine resolutions correspond to high frequencies, which is not strictly true. For instance, a point on a flat, monochromatic surface remains low frequency, irrespective of resolution. We will show in Section \ref{sec_ablation} that rendering improvements can be achieved just by removing frequency masking from INGP.

Secondly, INGP is fast-converging. Its aggressive optimization dynamics amplify the impact of even small adjustments. This is a challenge for model-based methods, which focus on tweaking the Vanilla MLP's architecture in NeRF to improve its few-shot capabilities \cite{mi-MLP}. We will again show in Section \ref{sec_ablation} that slight architectural changes, like simply adding \textit{one} more layer and some residual connections, degrade few-shot INGP's performance.

The above discussion yields a two-fold motivation. Firstly, given the proven effectiveness of frequency-based regularization, we need to devise a method that accomplishes the same idea, but compatible with INGP. To this end, we decompose the rendered and ground-truth scenes into levels of frequencies, using the Discrete Wavelet Transform. Their frequency-space differences are minimized by our novel Discrete Wavelet loss. By setting a higher weight to low-frequency discrepancies, we can \textit{explicitly} emphasize their learning right in the training objective. Secondly, given the proven effectiveness of model-based methods, we need to devise a method that does not degrade INGP, which is sensitive to architectural changes. Our focus is on modelling the color-density interactions. While better modelled with separate MLP branches, they still require some levels of interactions \cite{mi-MLP}. We call this the ``cross-branch interactions''. To this end, we propose using multi-head attention modules at the input and output levels. This does not incur architectural changes to the MLP, and does not degrade performance like other model-based methods.

In summary, our contributions are as follows:

\begin{itemize}
  \setlength\itemsep{0em}
  \item A unified framework, DWTNeRF, that blends the following prevalent approaches towards few-shot INGP:
  \item \textit{Regularization-based} approach with a novel Discrete Wavelet loss that explicitly decomposes the scenes into low and high-frequencies, emphasizing low frequencies at earlier training stages.
  \item \textit{Model-based} approach to enable color-density interactions using multi-head attention, which is compatible with fast-converging INGP;
  \item Through comprehensive experiments, we demonstrate that the proposed DWTNeRF is highly competitive with state-of-the-art methods, especially under highly challenging 2- to 4-shot conditions. The proposed DWTNeRF  is also much faster due to being INGP-based.
\end{itemize}
\section{Related Works}
\label{sec:relatedworks}

\subsection{Neural Radiance Fields}

\quad NeRF \cite{NeRF} has attained wide popularity in novel view synthesis and 3D scene representation tasks. It is self-supervised, able to construct 3D scenes without any 3D ground-truth data, relying only on multi-view images. Compared to earlier view synthesis methods, it is highly photorealistic. However, Vanilla NeRF is limited in practical applications \cite{NeRF-Review}. Therefore, further research has been aimed at faster training \& rendering \cite{PlenOctrees}, generalizability \cite{FeatureNeRF}, dynamic scenes \cite{DNeRF}, 3D generation \cite{CG-NeRF}, etc. In the sub-field of fast-training NeRF, INGP \cite{INGP} introduced a multi-resolution hash encoding whose CUDA implementation can render scenes within seconds. On top of INGP, we focus on another sub-field: few-shot NeRF, which aims to reduce its high dependence on a dense training set of multi-view images. 

\subsection{Few-shot NeRF} \label{sec_fewshot_related}

\quad \textbf{Prior-based methods}. Those enable few-shot NeRF by training a generalized model on a large dataset of diverse scenes or by incorporating pre-trained priors in their training objectives. pixelNeRF \cite{pixelNeRF} leveraged a feature encoder to extract image features from sparse input views, and conditioned a NeRF on these features. DietNeRF \cite{DietNeRF} introduced a semantic consistency loss that minimizes discrepancies in embeddings of rendered and ground-truth views, encouraging high-level semantic similarities. RegNeRF \cite{RegNeRF} leveraged a trained normalizing flow model to regularize color patches of unobserved views. SparseNeRF \cite{SparseNeRF} introduced a local depth ranking loss based on priors from a trained depth estimation model. SPARF \cite{SPARF} minimized coordinate-space discrepancies between pixel matches in multiple ground-truth views, predicted by a trained feature matching model. This encourages a global and geometrically accurate solution. GeCoNeRF \cite{GeCoNeRF} extended this concept of geometric consistency to the feature-space, regularizing both semantically and structurally.  

\textbf{Regularization-based methods}. Those enable few-shot NeRF by additionally introducing regularization terms, optimizing in a per-scene manner. InfoNeRF \cite{InfoNeRF} introduced a KL-divergence loss that enforces consistent density distributions across neighboring rays. RegNeRF \cite{RegNeRF} introduced a depth smoothness loss that penalizes depth discrepancies between neighboring points. Mip-NeRF 360 \cite{mipnerf360} introduced a distortion loss that reduces floating artifacts by minimizing the weighted distances between sampled points. DiffusioNeRF \cite{DiffusioNeRF} introduced a full geometry loss that encourages the weights of sampled points along each ray to sum to unity, which ensures the rays are absorbed fully by the scene's geometry. Another significant work is FreeNeRF \cite{FreeNeRF}, which did not introduce new terms but proposed masking input encodings in a coarse-to-fine manner, preventing NeRF from overfitting on high frequencies in early training stages. CombiNeRF \cite{CombiNeRF}, which our work is based on, synergistically combined all these techniques on top of INGP, achieving SOTA results.

\textbf{Model-based methods}. \cite{mi-MLP} shows that Vanilla MLP is not sufficient for few-shot NeRF. They introduced two main modules, which we term ``Residual Connections'' and ``Element-wise Cross-branch Interactions''. In the former, the encoded inputs are fed into each MLP layer, ensuring a shorter connection between the inputs and outputs. This is similar to ResNet \cite{ResNet} \& DenseNet's \cite{DenseNet} residual connections. In the latter, the colors \& densities are modeled using separate MLP branches, and cross-branch interactions are learned by a simple element-wise addition between corresponding layers. This pioneering work presented a new direction towards few-shot NeRF: adjusting Vanilla MLP's architecture. 

However, aside from DiffusioNeRF and CombiNeRF, none of the works above are based on INGP. Our experiments reveal that not all few-shot methods are transferable across Vanilla NeRF and INGP. Precisely, frequency regularization and model-based approaches are not trivially transferable. This is due to INGP's multi-resolution encoding and fast convergence, as explained in Section \ref{sec:intro}. To this end, we present a blend of regularization- and model-based approaches that are INGP-compatible.
\section{Preliminaries}

\subsection{Neural Radiance Fields}

\quad NeRF optimizes a 5D plenoptic function that represents a 3D volume, $f(\mathbf{x}, \mathbf{d})$, where $\mathbf{x} = (x, y, z)$ is a 3D spatial position viewed from a unit direction $\mathbf{d} = (\theta, \phi)$. $f$ outputs a view-dependent RGB color $\mathbf{c}$ and a differential volume density $\sigma$, parameterized by an MLP. To calculate the colors in a pixel grid, we start by shooting a ray $\mathbf{r}(t)$ into the 3D scene, through a pixel $\mathbf{p}$. The ray has an origin at $\mathbf{o}$ and a 3D direction $\mathbf{d}$. Along the ray, we sample multiple points. $f$ learns the color $\mathbf{c}$ and the density $\sigma$ of each point. The final color of the pixel $\mathbf{p}$, $\mathbf{c}(\mathbf{r})$, is calculated with simplified volume rendering \cite{VolumeRendering}:

\begin{equation} \label{eq_volumerender}
\mathbf{c}(\mathbf{r}) = \int_{t_n}^{t_f} T(t)\sigma(\mathbf{r}(t))\mathbf{c}(\mathbf{r}(t),\mathbf{d}) \,dt\
\end{equation}
where $t_n$ and $t_f$ denote the lower and higher-bound spatial positions of the sampled points. The color of a point $t$ is $\mathbf{c}(\mathbf{r}(t),\mathbf{d})$, weighted by the point's density $\sigma(\mathbf{r}(t))$. This is again weighted by the transmittance $T(t)$, the exponential of the negative integral of the differential density. It can be interpreted as the probability that the ray traverses uninterrupted from $t_n$ to $t$ \cite{rendering-book}. The process repeats for all rays forming a batch $\mathcal{R}$. NeRF is optimized by minimizing the MSE between predicted colors $\mathbf{c}(\mathbf{r})$ and ground-truth colors $\mathbf{c}_{gt} (\mathbf{r})$, under the same pose $\mathbf{p}$:


\begin{equation*}
    \mathcal{L}_{MSE} = \frac{1}{|\mathcal{R}|}\sum_{\mathbf{r} \in \mathcal{R}}||\mathbf{c}(\mathbf{r}) - \mathbf{c}_{gt} (\mathbf{r})||_2^2
\end{equation*}

\begin{figure*}[ht]
\vskip 0.2in
\begin{center}
\centerline{\includegraphics[width=\textwidth]{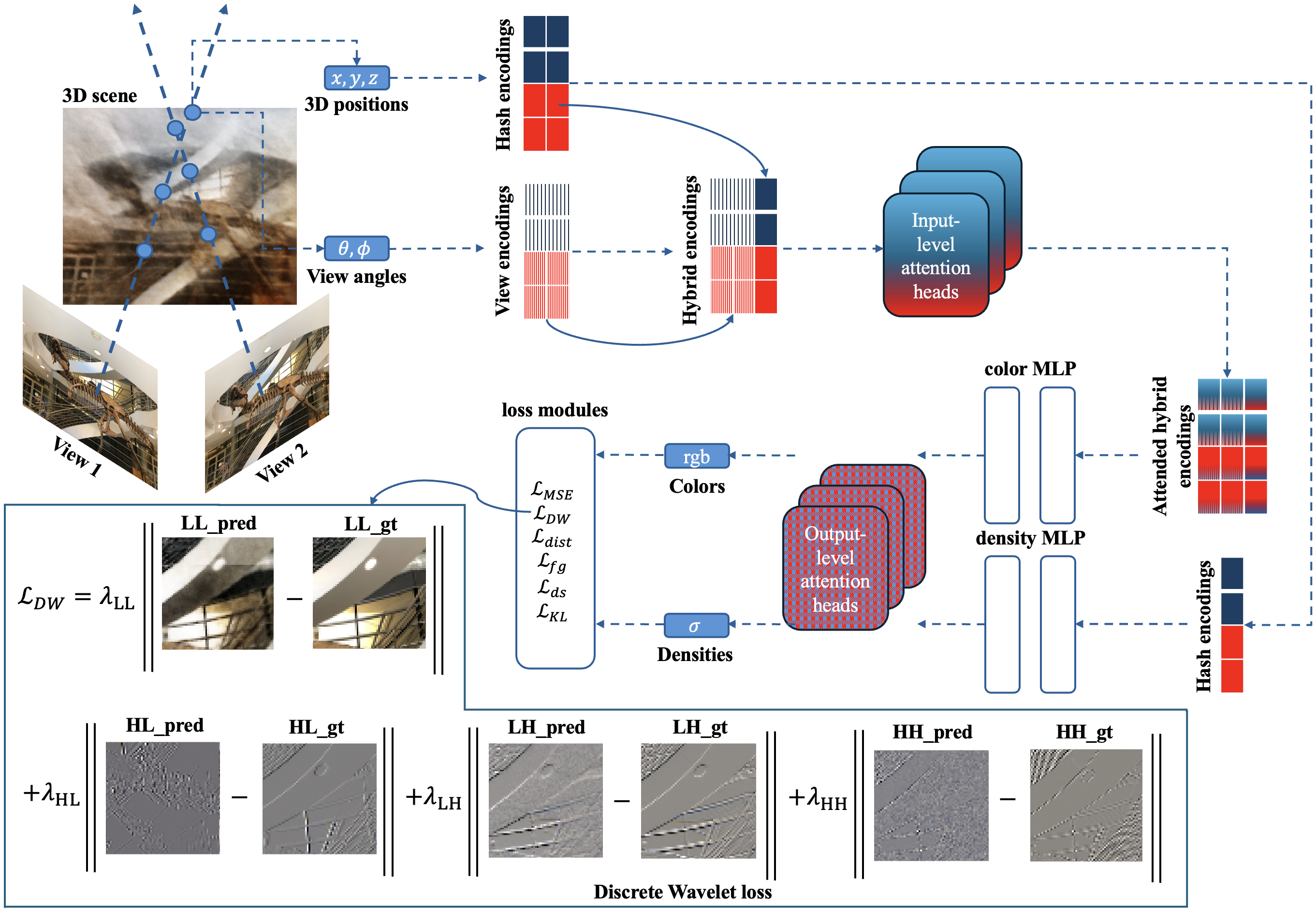}}
\caption{Architecture diagram of DWTNeRF. Firstly, we encode 3D positions with INGP's \cite{INGP} hash encoding, and the viewing directions with spherical harmonics \cite{Spherical}. Afterwards, we concatenate the view encodings with the hash encodings, to produce the ``hybrid encodings''. We facilitate cross-branch interactions by feeding them into a multi-head attention \cite{MHA} module. The attended hybrid encodings are inputs to the color MLP, while the hash encodings are inputs to the density MLP. Then, the MLP outputs from both branches are concatenated together, and again fed into another multi-head attention module. To supervise the quality of novel views, we use the usual photometric loss ($\mathcal{L}_{MSE}$) between the predicted and ground-truth views, as well as other regularization terms that are used in CombiNeRF \cite{CombiNeRF}.}
\label{fig_arch}
\end{center}
\vskip -0.2in
\end{figure*}

\subsection{Discrete Wavelet Transform (DWT)}


\quad Here, we explain the DWT, which our Discrete Wavelet loss (Section \ref{sec_DWloss}) is based on. Suppose that the ray batch $\mathcal{R}$ consists of adjacently sampled rays $\mathbf{r}$. This amounts to a 2D image $\mathbf{I}$, which can be compared with the ground-truth image $\mathbf{I}_{gt}$. We can perform the DWT on $\mathbf{I}$ to receive the four following sub-bands:

\begin{equation*}
    \mathbf{I}^{LL} = \mathbf{L}\mathbf{I}\mathbf{L}^T;     \mathbf{I}^{LH} = \mathbf{H}\mathbf{I}\mathbf{L}^T
\end{equation*}
\begin{equation*}
    \mathbf{I}^{HL} = \mathbf{L}\mathbf{I}\mathbf{H}^T;
    \mathbf{I}^{HH} = \mathbf{H}\mathbf{I}\mathbf{H}^T
\end{equation*}
where $\mathbf{L}$ and $\mathbf{H}$ are the low- and high-pass 2D matrices, respectively. $\mathbf{L}$ is constructed with shifting rows of its corresponding 1D low-pass wavelet filter $\bm{\ell}$, and similarly for $\mathbf{H}$. If $\bm{\ell}$ is orthogonal, then the 1D high-pass filter $\bm{h}$ is derived trivially from $\bm{\ell}$, and $\mathbf{H}$ trivially from $\mathbf{L}$ \cite{wavelet-book}. $\mathbf{I}^{LL}$ is called the LL sub-band of $\mathbf{I}$, and so on for other components. Intuitively, $\mathbf{I}^{LL}$ represents the low-frequency components of the image. $\mathbf{I}^{LH}$ and $\mathbf{I}^{HL}$ represent high frequencies in one direction, but low frequencies in another. Finally, $\mathbf{I}^{HH}$ represents the high frequencies in both vertical \& horizontal directions. Because $\mathbf{L}$ is constructed with shifting rows of $\bm{\ell}$ (and similarly for $\mathbf{H}$), the sub-bands are only a quarter of the original image. This is a concern in applying the DW loss, which we will explain later.
\section{Methodology}

\subsection{Overview}

\quad Figure \ref{fig_arch} shows the architecture diagram of our DWTNeRF. It consists of two main modules: the Discrete Wavelet (DW) loss and cross-branch interactions. The DW loss decomposes rendered and ground-truth views into levels of frequencies, and minimizes the differences in their frequency representations. By setting a higher weight for low-frequency differences, we can explicitly emphasize them in earlier training stages, thereby reducing overfitting on high frequencies. This represents a \textit{regularization-based} approach that is frequency-centric. On the other hand, the cross-branch modules utilize multi-head attention to capture interactions not only between colors and densities but also among neighboring 3D points. This helps mitigate overfitting to high-frequency details by constraining color predictions through density-based, position-aware relationships. This also represents a \textit{model-based} method that is compatible with INGP, which is highly sensitive to architectural changes.

\subsection{Scene Decomposition with Discrete Wavelet loss} \label{sec_DWloss}

\quad Given a pair of predicted view $\mathbf{I}$ and ground-truth view $\mathbf{I}_{gt}$, the Discrete Wavelet (DW) loss minimizes the differences of their frequency representations:

\begin{equation*}
    \mathcal{L}_{DW} = \sum_{sb \in\{LL, LH, HL, HH\}}\lambda_{sb}||\mathbf{I}^{sb}-\mathbf{I}_{gt}^{sb}||_2^2
\end{equation*}
where $\lambda_{sb}$ is the weighting for the corresponding sub-bands. By setting $\lambda_{LL}$ higher than other sub-bands, we have explicitly prioritized learning of low frequencies. The DW loss is incorporated with the photometric loss $\mathcal{L}_{MSE}$, as well as all regularization losses used in CombiNeRF \cite{CombiNeRF}:

\begin{equation*}
\begin{split}
    \mathcal{L} = \mathcal{L}_{MSE} + \mathcal{L}_{DW} + \lambda_{dist}\mathcal{L}_{dist} + \lambda_{fg}\mathcal{L}_{fg} + \\ \lambda_{ds}\mathcal{L}_{ds} + \lambda_{KL}\mathcal{L}_{KL}
\end{split}
\end{equation*}
where $\mathcal{L}_{dist}$ is the distortion loss, introduced by Mip-NeRF 360 \cite{mipnerf360}. $\mathcal{L}_{fg}$ is the full geometry loss, introduced by DiffusioNeRF \cite{DiffusioNeRF}. $\mathcal{L}_{ds}$ is the depth smoothness loss, introduced by RegNeRF \cite{RegNeRF}. $\mathcal{L}_{KL}$ is the KL divergence loss, introduced by InfoNeRF \cite{InfoNeRF}. The DW loss allows us to emphasize low frequencies directly in the training objective.

\begin{figure}[ht] 
\vskip 0.2in
\begin{center}
\centerline{\includegraphics[width=0.9\columnwidth]{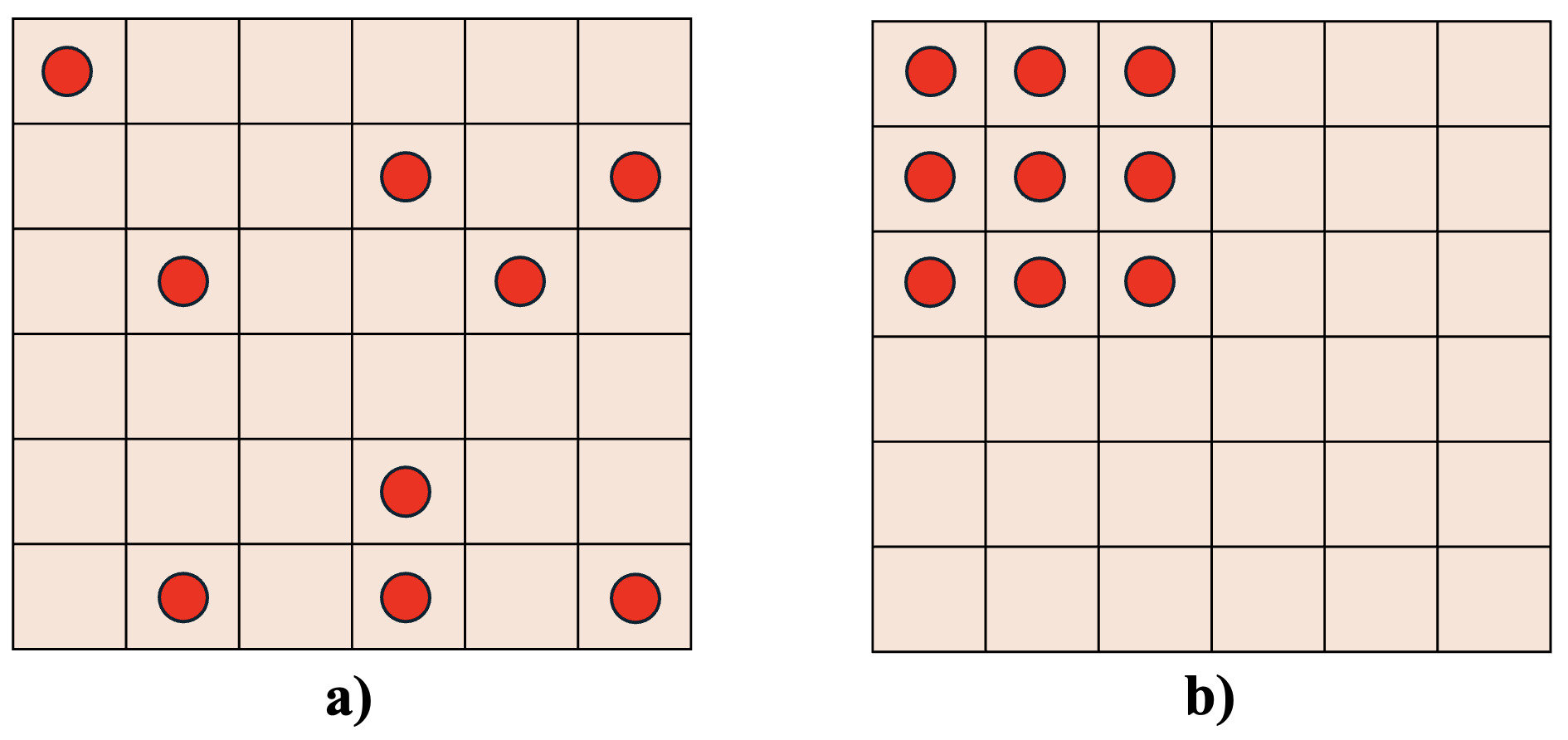}}
\caption{Random (left) and patch-based ray sampling (right). Patch sampling immediately forms a 2D image, and is compatible with the Discrete Wavelet Transform.}
\label{fig_patch}
\end{center}
\vskip -0.2in
\end{figure}

\textbf{Patch-based Rendering}. The DW loss requires adjacently sampled rays $\mathbf{r}$ that make up a 2D image $\mathbf{I}$. This requires patch-based ray sampling. As illustrated in Figure \ref{fig_patch}b, rays sampled in patches are adjacent to each other, making up a small square grid. Figure \ref{fig_patch}a shows rays sampled randomly throughout the image grid. We use random ray sampling for calculating $\mathcal{L}_{MSE}$, and patch-based for $\mathcal{L}_{DW}$. This design choice is intentional: it considers both global structures (random rays) and local structures (patch-based rays) in one iteration.

\textbf{Lazy Regularization}. The sub-bands are 4 times smaller than the original image $\mathbf{I}$. For the DW loss to be feasible, however, the sub-bands still need to be visually meaningful. For example, $\mathbf{I}^{LH}$ and $\mathbf{I}^{HL}$ can highlight vertical and horizontal edges of a scene. By minimizing their differences with the ground-truths, we can better preserve those mid-frequency-level structures. This requires the rendered image $\mathbf{I}$ to be large in the first place, or else its sub-bands would be too coarse. This incurs heavy computational overhead, because an $N$-sized square patch requires $N^2$ number of rays. We alleviate the problem as follows. Firstly, we follow the lazy regularization procedure inspired by \cite{StyleGAN}, where we only calculates the DW loss every few iterations. Secondly, since we only prioritize low frequencies earlier in the training, we can safely stop calculating the DW loss later on. We present the full training algorithm of DWTNeRF at Algorithm \labelcref{alg:dwtnerf}. Here, we only detail $\mathcal{L}_{MSE}$ and $\mathcal{L}_{DW}$ for simplicity.

\begin{algorithm}[tb]
   \caption{DWTNeRF Training Algorithm}
   \label{alg:dwtnerf}
\begin{algorithmic}
   \STATE {\bfseries Input:} Dataset of observed views $\mathcal{D}=\{(\mathbf{I}_{full}, \mathbf{p})\}$, with pose $\mathbf{p}$ for each full-resolution image $\mathbf{I}_{full}$; current iteration $t$; maximum iteration $T$; randomly sampled batch size $|\mathcal{R}|$; patch-based sampled batch size $|\mathcal{R}_P|$; DW loss's interval $K_{DW}$; DW loss's maximum iteration $T_{DW}$; sub-band weights $\lambda_{(\cdot)}$; learning rate $\eta_t$
   \STATE {\bfseries Output:} Trained NeRF $f_{\boldsymbol{\theta}}(\cdot,\cdot)$
   \STATE Initialize NeRF $f_{\boldsymbol{\theta}}(\cdot,\cdot)$
   \REPEAT
   \STATE Randomly sample a ray batch $\mathcal{R}$, with corresponding ground-truth colors $\mathbf{c}_{gt}(\mathcal{R})$
   \STATE Compute predicted colors $\mathbf{c}(\mathcal{R})$ by Equation \eqref{eq_volumerender}
    \STATE $\mathcal{L} \gets \mathcal{L}_{MSE}(\mathbf{c}(\mathcal{R}), \mathbf{c}_{gt}(\mathcal{R}))$
   \IF{$t \bmod K_{DW} = 0$ \AND $t < T_{DW}$} 
   \STATE Sample a patch-based ray batch $\mathcal{R}_P$, with corresponding ground-truth colors $\mathbf{c}_{gt}(\mathcal{R}_P) \equiv \mathbf{I}_{gt}$
   \STATE Compute predicted colors $\mathbf{c}(\mathcal{R}_P) \equiv \mathbf{I}$ by Equation \eqref{eq_volumerender}
   \STATE Compute predicted sub-bands $\mathbf{I}^{LL}$, $\mathbf{I}^{LH}$, $\mathbf{I}^{HL}$, $\mathbf{I}^{HH}$
   \STATE Compute ground-truth sub-bands $\mathbf{I}_{gt}^{LL}$, $\mathbf{I}_{gt}^{LH}$, $\mathbf{I}_{gt}^{HL}$, $\mathbf{I}_{gt}^{HH}$
    \STATE $\mathcal{L} \gets \mathcal{L} +\mathcal{L}_{DW}(\mathbf{I}^{(\cdot)}, \mathbf{I}_{gt}^{(\cdot)})$
   \ENDIF
   \STATE $\boldsymbol{\theta} \gets \texttt{Adam}(\boldsymbol{\theta}, \eta_t, \nabla_{\boldsymbol{\theta}}\mathcal{L})$
   \UNTIL{$t > T$}
\end{algorithmic}
\end{algorithm}

\subsection{Cross-branch Interactions with Multi-head Attention} \label{sec_MHA}

\quad Our approach for cross-branch interactions consists of the following:

\textbf{I. Cross-branch Concatenation}. We are given the hash encoding $\gamma_{0:N_{\mathbf{x}}}(\mathbf{x})$ and the view encoding $\gamma_{0:N_{\mathbf{d}}}(\mathbf{d})$. Here, the subscripts denote the dimension of the encoding. $0:N_{(\cdot)}$ means the full dimension. $N_{\mathbf{x}}$ and $N_{\mathbf{d}}$ represent the highest dimensions of the hash and view encodings. The first stage involves concatenating the view encoding with all-but-the-first-dimension of the hash encoding. This can be written as $\gamma'(\mathbf{d}, \mathbf{x}) = \left[ \gamma(\mathbf{d}), \gamma_{1:N_{\mathbf{x}}}(\mathbf{x}) \right]$, where $[\cdot,\cdot]$ denotes the concatenation operator. We call this the ``hybrid encoding''. The hash encoding only retains its first dimension: $\gamma'(\mathbf{x})=\gamma_{0}(\mathbf{x})$. The intuition is as follows. We will use $\gamma'(\mathbf{x})$ as inputs to the density MLP, and $\gamma'(\mathbf{d},\mathbf{x})$ as inputs to the color MLP. Densities are irrespective of viewing directions and should only be position-dependent, so we make $\gamma'(\mathbf{x})$ extremely lean to avoid over-parameterizing them. On the other hand, colors are both view- and position-dependent, which explains the concatenation.

\textbf{II. Input-level Multi-head Attention}. We now have the ``trimmed'' hash encoding $\gamma'(\mathbf{x})$ and hybrid encoding $\gamma'(\mathbf{d},\mathbf{x})$. We push $\gamma'(\mathbf{d},\mathbf{x})$ into a multi-head attention module to learn the interactions between $\gamma(\mathbf{d})$ and $\gamma_{1:N_{\mathbf{x}}}(\mathbf{x})$. Again, we leave $\gamma'(\mathbf{x})$ intact to not over-parameterize it. The multi-head attention is applied as: $\texttt{MultiHead}(\mathbf{Q}, \mathbf{K}, \mathbf{V}) = [\texttt{head}_0,...,\texttt{head}_H]\mathbf{W}^O$, where $\texttt{head}_i = \texttt{Attn}(\mathbf{Q}\mathbf{W}_i^{\mathbf{Q}},\mathbf{K}\mathbf{W}_i^{\mathbf{K}}, \mathbf{V}\mathbf{W}_i^{\mathbf{V}})$. Here, \texttt{Attn} is simply the attention layer \cite{MHA}. $\mathbf{Q}$, $\mathbf{K}$, $\mathbf{V}$ are the query, key and value matrices, all of which are set to $\gamma'(\mathbf{d},\mathbf{x})$. $\mathbf{W}_i^{\mathbf{Q}}$, $\mathbf{W}_i^{\mathbf{K}}$, $\mathbf{W}_i^{\mathbf{V}}$ are the corresponding learnable projection matrices. We note that $\gamma'(\mathbf{d},\mathbf{x})$ is a tensor of size $|\mathcal{R}| \times (N_{\mathbf{x}}-1+N_{\mathbf{d}})$. The choice of using attention is deliberate. Along the vertical dimension, it learns the interactions between all rays $\mathbf{r}$ in the batch $\mathcal{R}$. This equates to learning the interactions between neighboring 3D points. Along the horizontal dimension, it learns the interactions between the positions and viewing directions. Using many heads, up to a maximum $H$, provides many of such interactions. The heads are concatenated, and then projected by another learnable matrix $\mathbf{W}^O$ to ensure outputs and inputs are of the same size. We now write the attended hybrid encoding as $\gamma''(\mathbf{d}, \mathbf{x})$. The plenoptic function is now $f(\gamma'(\mathbf{x}),\gamma''(\mathbf{d},\mathbf{x}))=\sigma, \mathbf{c}$, parameterized by two separate MLP branches for densities and colors.

\textbf{III. Output-level Multi-head Attention}. $f$'s outputs are the ``preliminary'' colors $\mathbf{c}$ and densities $\sigma$. We again learn their interactions. We concatenate them together, and feed them into a multi-head attention module. Here, $\mathbf{Q}$, $\mathbf{K}$, $\mathbf{V}$ are set to $\boldsymbol{\zeta}=[\sigma, \mathbf{c}]$. All intuitions in the second stage still hold. We ``un-concatenate'' the attended $\boldsymbol{\zeta}'$ to obtain the final predictions. Specifically, the densities $\sigma'$ are the first dimension of $\boldsymbol{\zeta}'$, and the colors $\mathbf{c}'$ are the three remaining dimensions.

\section{Experiments}

\subsection{Datasets \& Implementation Details}

\quad \textbf{Datasets \& Metrics}. DWTNeRF is evaluated on the LLFF \cite{LLFF} dataset and the NeRF-Synthetic dataset \cite{NeRF}, under few-shot conditions. Those are popular benchmarks for novel view synthesis. For the LLFF dataset, we evaluated the test metrics under 2/3/6/9 input views. For the NeRF-Synthetic (NS) dataset, we trained on 4 views and tested on 25 views. Those protocols are similar to previous works. We employed the PSNR, SSIM \cite{SSIM} and LPIPS \cite{LPIPS} metrics to evaluate the quality of novel views.

\begin{figure}[ht]
\vskip 0.2in
\begin{center}
\centerline{\includegraphics[width=\columnwidth]{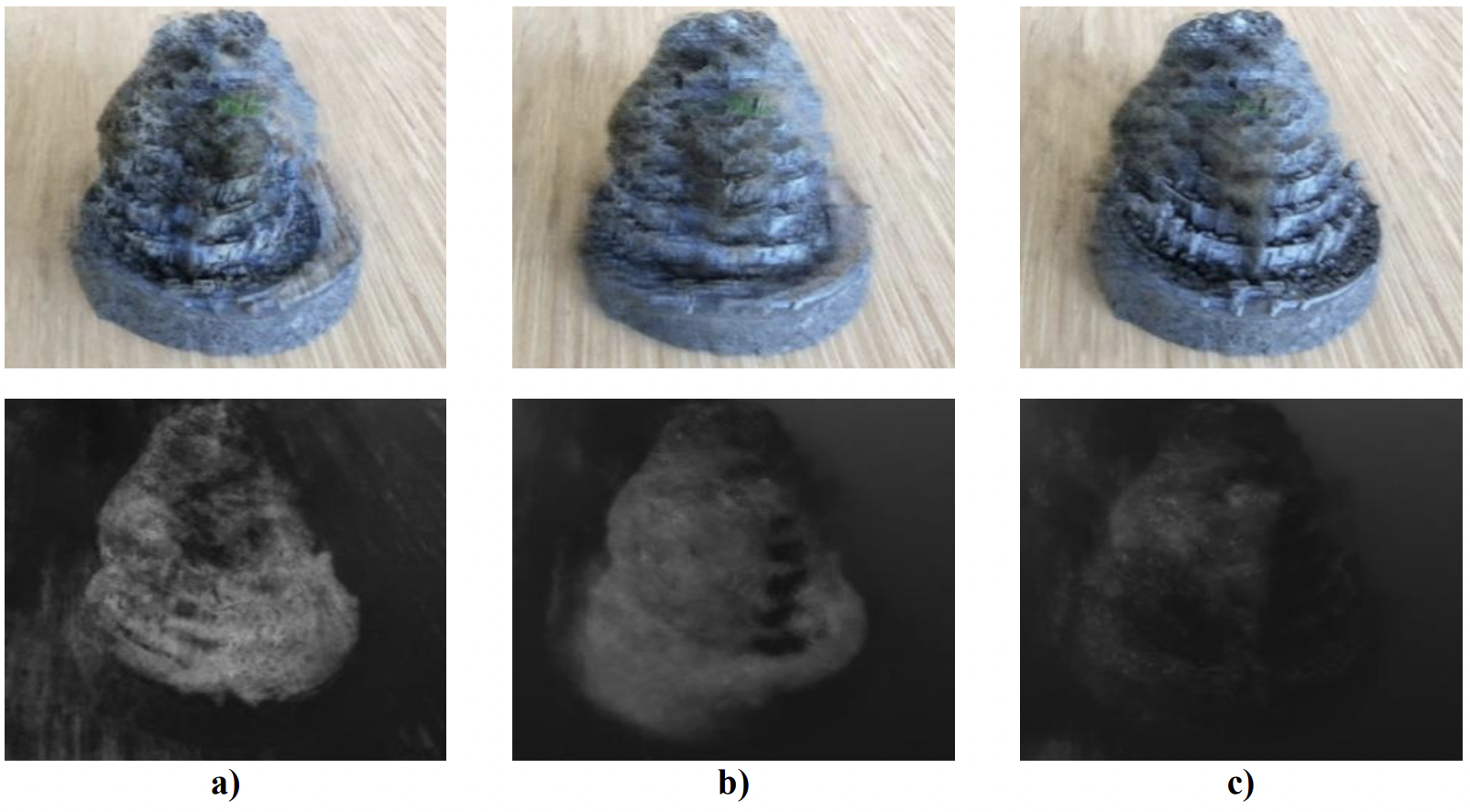}}
\caption{Qualitative results for Vanilla INGP (a), CombiNeRF (b) and DWTNeRF (c) on the LLFF \cite{LLFF} dataset (``fortress'' scene). Top row depicts the novel views, while bottom row depicts the corresponding depth maps. Our DWTNeRF retains finer details better. The depth map generated by DWTNeRF is also more factual.}
\label{fig_results_vis_LLFF}
\end{center}
\vskip -0.2in
\end{figure}

\begin{figure}[ht]
\vskip 0.2in
\begin{center}
\centerline{\includegraphics[width=0.75\columnwidth]{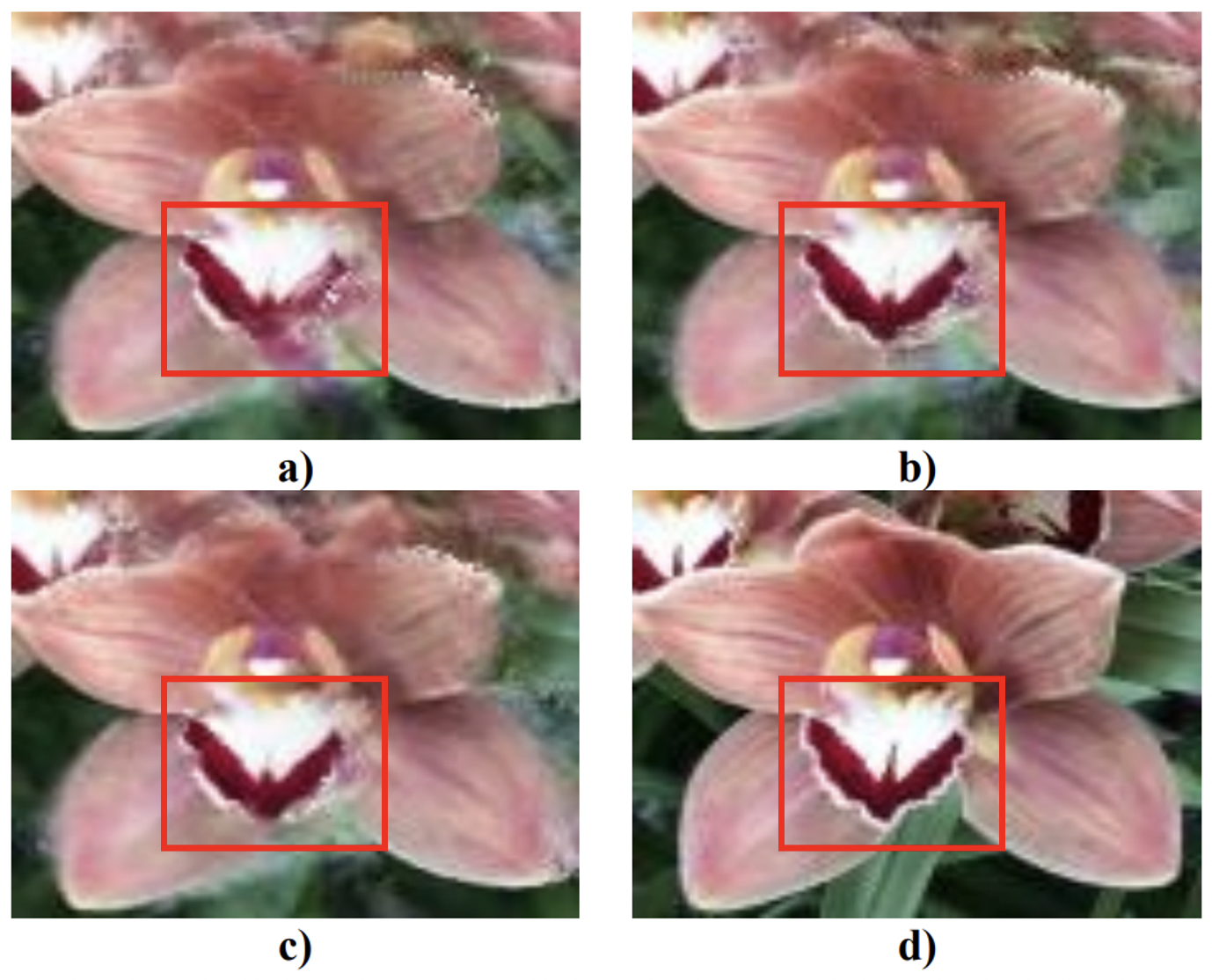}}
\caption{Qualitative results for CombiNeRF (a), DWTNeRF with no DW loss (b), DWTNeRF with both modules (c) and ground-truth (d) on the LLFF \cite{LLFF} dataset (``orchids'' scene). Using the DW loss preserves structures better, which in turn produces the least pink-colored ``hallucinations'', as depicted in the \textbf{\textcolor{Red}{red}} region.}
\label{fig_results_vis_abs_LLFF}
\end{center}
\vskip -0.2in
\end{figure}

\textbf{Implementation Details}. DWTNeRF is built on top of CombiNeRF \cite{CombiNeRF}, which in turn is built on a Pytorch implementation of INGP, torch-ngp \cite{torch-ngp}. We kept all default settings of CombiNeRF, and trained for 10K iterations in all experiments. We randomly sampled 4096 rays (for LLFF) and 7008 rays (for NS) in each iteration. For DW loss, we additionally sampled 36864 rays, leading to 192 $\times$ 192 square patches. We calculated the DW loss every 10 (for LLFF) and 150 (for NS) iterations, but stopped altogether at 5K iterations to facilitate high-frequency learning. For LLFF, we set the LL sub-band weight as 0.4, the other sub-bands were 0.2. For NS, LL sub-band weight was 0.04, the rest were all 0.02. This ensures that the low frequencies were the most important. Finally, we used 2 (for LLFF) and 1 (for NS) attention heads for both input and output levels. All trainings were done with GeForce RTX 4070 Ti Super.

\subsection{Qualitative Results}

\begin{table*}[t]
\caption{Comparison of DWTNeRF against SOTA methods under 2/3/6/9 views, LLFF dataset \cite{LLFF}}
\label{tab_SOTA_LLFF}
\vskip 0.15in
\begin{center}
\begin{small}
\begin{sc}
\resizebox{\textwidth}{!}{%
\begin{tabular}{lccccccccccccc}
\toprule
Method & Venue & \multicolumn{4}{c}{PSNR $(\uparrow)$} & \multicolumn{4}{c}{SSIM $(\uparrow)$} & \multicolumn{4}{c}{LPIPS $(\downarrow)$} \\
\cmidrule(lr){3-6} \cmidrule(lr){7-10} \cmidrule(lr){11-14}
 & & 2-view & 3-view & 6-view & 9-view & 2-view & 3-view & 6-view & 9-view & 2-view & 3-view & 6-view & 9-view \\
\midrule
Vanilla INGP & arXiv `22 & 16.59 & 17.71 & 22.03 & 24.21 & 0.458 & 0.544 & 0.738 & 0.811 & 0.359 & 0.303 & 0.149 & 0.101 \\
GeCoNeRF & ICML `23 & - & 18.77 & - & - & - & 0.596 & - & - & - & 0.338 & - & - \\
FreeNeRF & CVPR `23 & - & 19.63 & 23.73 & 25.13 & - & 0.612 & 0.779 & 0.827 & - & 0.308 & 0.195 & 0.160 \\
DiffusioNeRF & CVPR `23 & - & 19.79 & 23.79 & 25.02 & - & 0.568 & 0.747 & 0.785 & - & 0.338 & 0.237 & 0.212 \\
mi-MLP & CVPR `24 & - & 19.75 & 23.57 & 25.15 & - & 0.614 & 0.788 & 0.834 & - & 0.300 & 0.163 & 0.140 \\
CombiNeRF $\dagger$ & 3DV `24 & 16.47 & 20.12 & \textbf{24.05} & \textbf{25.24} & 0.475 & 0.676 & \textbf{0.801} & \textbf{0.836} & 0.336 & 0.197 & 0.105 & \textbf{0.084} \\
DWTNeRF (ours) & - & \textbf{16.73} & \textbf{20.38} & 24.04 & 25.20 & \textbf{0.479} & \textbf{0.677} & \textbf{0.801} & 0.835 & \textbf{0.322} & \textbf{0.193} & \textbf{0.104} & \textbf{0.084} \\
\bottomrule
\end{tabular}%
}
\end{sc}
\end{small}
\end{center}
\vskip -0.1in
\end{table*}

\quad Figure \ref{fig_results_vis_LLFF} shows the qualitative results between Vanilla INGP, CombiNeRF and our DWTNeRF. The chosen scene is ``fortress''. Our DWTNeRF clearly better retains the fine structures of the fortress. As for depth maps, both CombiNeRF and DWTNeRF dramatically lessen floating artifacts that are extremely prevalent in Vanilla INGP. DWTNeRF's depth map is also more factual than CombiNeRF: the fortress should generally be as far to the camera as the table. 

Figure \ref{fig_results_vis_abs_LLFF} shows a more ablation-level analysis that demonstrates the effects of using the DW loss. The chosen scene is ``orchids''. With the DW loss, we additionally gain a supervision on the low- and middle frequencies. This translates to better preserved structures, and less 3D ``hallucinations''. We provide more qualitative results of both LLFF and NeRF-Synthetic at Section \ref{app_sec_more_vis} of Appendix.

\subsection{Quantitative Results}

\begin{table}[t]
\caption{Comparison of DWTNeRF against SOTA methods under 4 views, NeRF-Synthetic dataset \cite{NeRF}}
\label{tab_SOTA_NS}
\vskip 0.15in
\begin{center}
\begin{small}
\resizebox{\columnwidth}{!}{%
\begin{sc}
\begin{tabular}{lcccr}
\toprule
Method & Venue & PSNR ($\uparrow$) & SSIM ($\uparrow$) & LPIPS ($\downarrow$) \\
\midrule
DietNeRF & ICCV `21 & 15.42 & 0.730 & 0.314 \\
Vanilla INGP $\dagger$ & arXiv `22 & 17.49 & 0.734 & 0.357 \\
RegNeRF & CVPR `22 & 13.71 & 0.786 & 0.346 \\
InfoNeRF & CVPR `22 & 18.44 & 0.792 & 0.223 \\
CombiNeRF $\dagger$ & 3DV `24 & 19.15 & 0.792 & 0.224 \\
DWTNeRF (Ours) & - & \textbf{19.25} & \textbf{0.793} & \textbf{0.222} \\
\bottomrule
\end{tabular}
\end{sc}
}
\end{small}
\end{center}
\vskip -0.1in
\end{table}

\quad \textbf{On LLFF}. We compared DWTNeRF against recent few-shot works published at top venues. Regularization-based methods include InfoNeRF \cite{InfoNeRF}, FreeNeRF \cite{FreeNeRF} and CombiNeRF \cite{CombiNeRF}. Prior-based methods include DietNeRF \cite{DietNeRF}, RegNeRF \cite{RegNeRF}, DiffusioNeRF \cite{DiffusioNeRF} and GeCoNeRF \cite{GeCoNeRF}. Model-based method includes mi-MLP \cite{mi-MLP}. Table \ref{tab_SOTA_LLFF} shows the quantitative results. Methods with the symbol $\dagger$ were reproduced on our machine, keeping all default settings. Unavailable metrics are denoted ``$-$''. The best-performing metrics under each few-shot condition are \textbf{bolded}. In the 2-shot case, DWTNeRF outperforms CombiNeRF by 1.6\% in PSNR, 0.84\% in SSIM and 4.17\% in LPIPS. The effects of our DWTNeRF are still strongly seen with 3-shot, but generally fade with higher views, suggesting its high compatibility with extreme few-shot conditions.

\textbf{On NeRF-Synthetic}. Table \ref{tab_SOTA_NS} shows the quantitative results. In the 4-shot case, DWTNeRF outperforms CombiNeRF by 0.5\% in PSNR, 0.1\% in SSIM and 0.9\% in LPIPS.

\subsection{Ablation Studies} \label{sec_ablation}

\quad In this section, we validated the design choices of DWTNeRF. Ablations were done on the 3-shot LLFF benchmark.

\textbf{On main modules}. Those are the DW loss and the cross-banch interactions. Table \ref{tab:module_abs} shows the effects of each module. The second row removes the frequency masking from CombiNeRF, given its incompatibility with INGP. This provides improvements in all metrics. The third row is our DWTNeRF, but with the DW loss only. Starting from the fourth row, we examine the effects of using multi-head attention in cross-branch interactions. The fourth row is DWTNeRF with only input-level attention. The fifth row only has output-level attention. The sixth row combines both levels. This produces stronger results over the previous two. The final row combines both DW loss and cross-branch interactions, which provides the strongest results.

\begin{table}[t]
\caption{Ablation on DWTNeRF's main modules, LLFF dataset \cite{LLFF}}
\label{tab:module_abs}
\vskip 0.15in
\begin{center}
\begin{small}
\resizebox{\columnwidth}{!}{%
\begin{sc}
\begin{tabular}{lcccr}
\toprule
Method & PSNR ($\uparrow$) & SSIM ($\uparrow$) & LPIPS ($\downarrow$) \\
\midrule
CombiNeRF (baseline) $\dagger$ & 20.12 & 0.676 & 0.197 \\
CombiNeRF (-freq. mask.) $\dagger$ & 20.19 & \textbf{0.677} & 0.194 \\
\hline
DWTNeRF (+dw.) & 20.25 & \textbf{0.677} & \textbf{0.193} \\
\hline
DWTNeRF (+inp. attn.) & 20.33 & 0.670 & 0.195 \\
DWTNeRF (+outp. attn.) & 20.34 & 0.670 & 0.195 \\
DWTNeRF (+full cb.) & 20.36 & 0.670 & 0.195 \\
\hline
DWTNeRF (+both) & \textbf{20.38} & \textbf{0.677} & \textbf{0.193} \\
\bottomrule
\end{tabular}
\end{sc}
}
\end{small}
\end{center}
\vskip -0.1in
\end{table}

\begin{table}[t]
\caption{Comparison against model-based methods, LLFF dataset \cite{LLFF}}
\label{tab:prior_and_model_abs}
\vskip 0.15in
\begin{center}
\begin{small}
\resizebox{\columnwidth}{!}{%
\begin{sc}
\begin{tabular}{lccccr}
\toprule
Method & Type & PSNR ($\uparrow$) & SSIM ($\uparrow$) & LPIPS ($\downarrow$) \\
\midrule
CombiNeRF (baseline) $\dagger$ & R & 20.12 & 0.676 & 0.197 \\
CombiNeRF (+res.) $\dagger$ & M, R & 20.20 & 0.664 & 0.198 \\
CombiNeRF (+res.+) $\dagger$ & M, R & 19.99 & 0.650 & 0.204 \\
CombiNeRF (+ele. cb.) $\dagger$ & M, R & 19.90 & 0.667 & 0.205 \\
DWTNeRF (+full cb.) & M, R & \textbf{20.36} & 0.670 & \textbf{0.195} \\
\bottomrule
\end{tabular}
\end{sc}
}
\end{small}
\end{center}
\vskip -0.1in
\end{table}

\begin{figure}[ht]
\vskip 0.2in
\begin{center}
\centerline{\includegraphics[width=0.8\columnwidth]{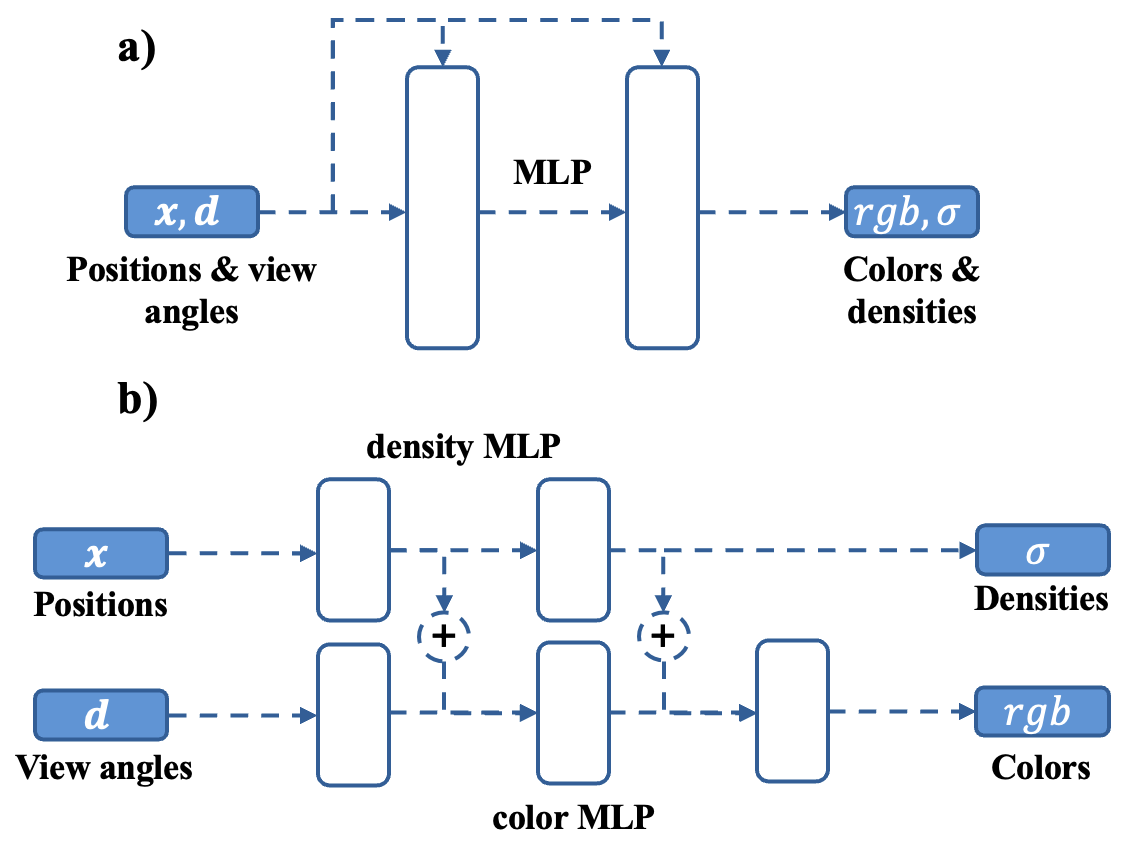}}
\caption{mi-MLP's \cite{mi-MLP} model-based methods: ``Residual Connections'' (a) and ``Element-wise Cross-branch Interactions'' (b)}
\label{arch_mimlp}
\end{center}
\vskip -0.2in
\end{figure}

\textbf{On model-based methods} We implemented two model-based approaches on top of CombiNeRF: ``Residual Connections'' (Figure \ref{arch_mimlp}a) and ``Element-wise Cross-branch Interactions'' (Figure \ref{arch_mimlp}b) - both introduced by mi-MLP \cite{mi-MLP}. Table \ref{tab:prior_and_model_abs} shows the results. The second row introduces residuals. The third row is the same, but we added one MLP layer, hoping to better highlight the residuals' effects. However, this degraded performances noticeably, highlighting the vulnerability of INGP to architectural changes. The fourth row shows mi-MLP's element-wise cross-branch interactions. Only our version of interactions, based on attention, is competitive with the baseline, and exceeds in PSNR and LPIPS. For more implementation details, please refer to Section \ref{app_sec_model} of Appendix.

\textbf{On the choice of wavelets}. We compared the effects of different Daubechies wavelets \cite{wavelet-db} on the DW loss. We started with the Haar ($n=1$) and then higher-order wavelets, up to the db3 ($n=3$). Generally, higher-order wavelets have more supports, and spreads the signal energy more evenly across sub-bands. Section \ref{app_sec_wavelet_coeffs} of Appendix shows the exact wavelet coefficients. Figure \ref{results_wavelets_small} compares scene decomposition using the Haar versus db2 ($n=2$). Compared to the ground-truths, the rendered sub-bands are much noisier, highlighting the purpose of the DW loss. The db2 provides finer details. However, table \ref{tab:wavelet_abs} shows no significant differences between the wavelets. This confirms our intuition that low-frequency supervision is sufficient.

\begin{table}[t]
\caption{Ablation on wavelets for DW loss}
\label{tab:wavelet_abs}
\vskip 0.15in
\begin{center}
\begin{small}
\resizebox{\columnwidth}{!}{%
\begin{sc}
\begin{tabular}{lcccr}
\toprule
Method & PSNR ($\uparrow$) & SSIM ($\uparrow$) & LPIPS ($\downarrow$) \\
\midrule
CombiNeRF (baseline) $\dagger$ & 20.12 & 0.676 & 0.197 \\
DWTNeRF (+dw. / haar) & \textbf{20.25} & 0.677 & 0.193 \\
DWTNeRF (+dw. / db2) & 20.23 & \textbf{0.678} & 0.193 \\
DWTNeRF (+dw. / db3) & 20.23 & \textbf{0.678} & \textbf{0.192} \\
\bottomrule
\end{tabular}
\end{sc}
}
\end{small}
\end{center}
\vskip -0.1in
\end{table}

\begin{figure}[ht]
\vskip 0.2in
\begin{center}
\centerline{\includegraphics[width=\columnwidth]{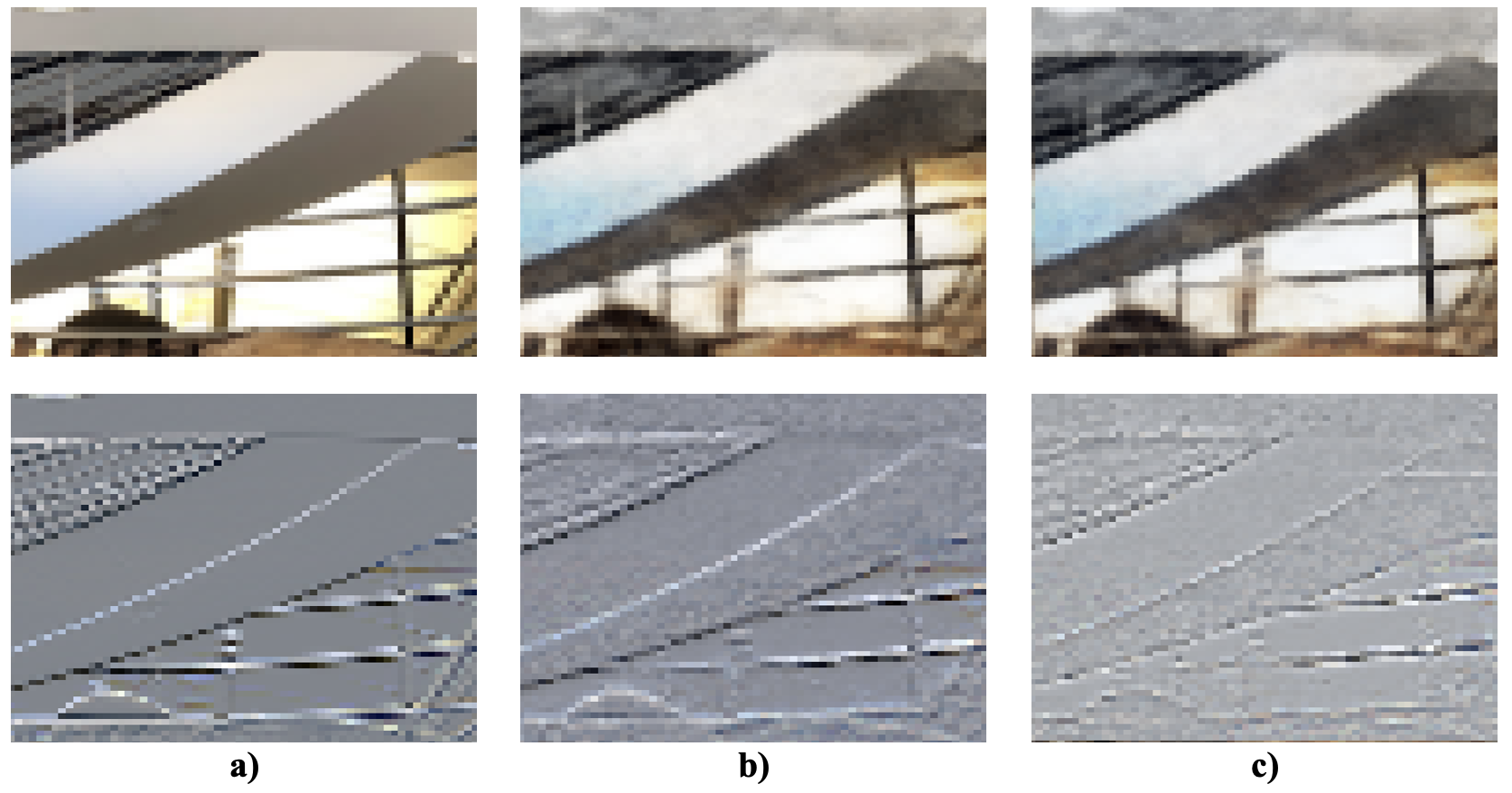}}
\caption{Decomposition using the Haar (a \& b) and db2 (c). We only show the LL (top) and LH (bottom) sub-bands. Column a) is the ground-truth LL, much less noiser than rendered LL at column b). Column c) uses higher-order wavelet, and shows finer details.}
\label{results_wavelets_small}
\end{center}
\vskip -0.2in
\end{figure}

\section{Conclusion}

\quad INGP-based models have distinct characteristics that might make them incompatible with recent few-shot approaches. Firstly, they are fast-converging, and vulnerable to architectural changes under few-shot conditions. This requires a re-thinking in \textit{model-based} approaches. For this, we introduce a cross-branch interaction pipeline based on multi-head attention, which acts on NeRF's inputs \& outputs, but not on the MLP layers. Secondly, based on multi-resolution, INGP does not explicitly map the inputs into a range of frequencies. This requires a re-thinking in \textit{frequency-} and \textit{regularization-based} approaches. For this, we introduce a novel Discrete Wavelet loss that decomposes the scenes into different frequencies and allows explicit prioritization of low frequencies in earlier training stages.

\textbf{Limitations \& Future Works}. Firstly, the DWT does not disentangle high-frequency components. For example, besides fine structural details, high frequencies can come from specularity and other light-dependent effects. Future works might explore scene decomposition at such further levels. Secondly, since the DW loss is used as an auxiliary loss alongside the photometric MSE loss, which compares pixel-wise differences regardless of frequencies, we believe some high-frequency overfitting still exists. Our DW loss \textit{emphasizes} low frequencies, but not \textit{eliminate} high frequencies in early training. Future works could tackle this problem, but interfering with the photometric loss can be tricky. We can also investigate the applicability of our methods to other representations, like SDF \cite{SDF, splatsdf} or 3DGS \cite{3DGS}.
{
    \small
    \bibliographystyle{ieeenat_fullname}
    \bibliography{main}
}
\twocolumn
\clearpage
\setcounter{page}{1}
\maketitlesupplementary

\section{More Visualizations} \label{app_sec_more_vis}

\begin{figure*}[ht]
\vskip 0.2in
\begin{center}
\centerline{\includegraphics[width=\textwidth]{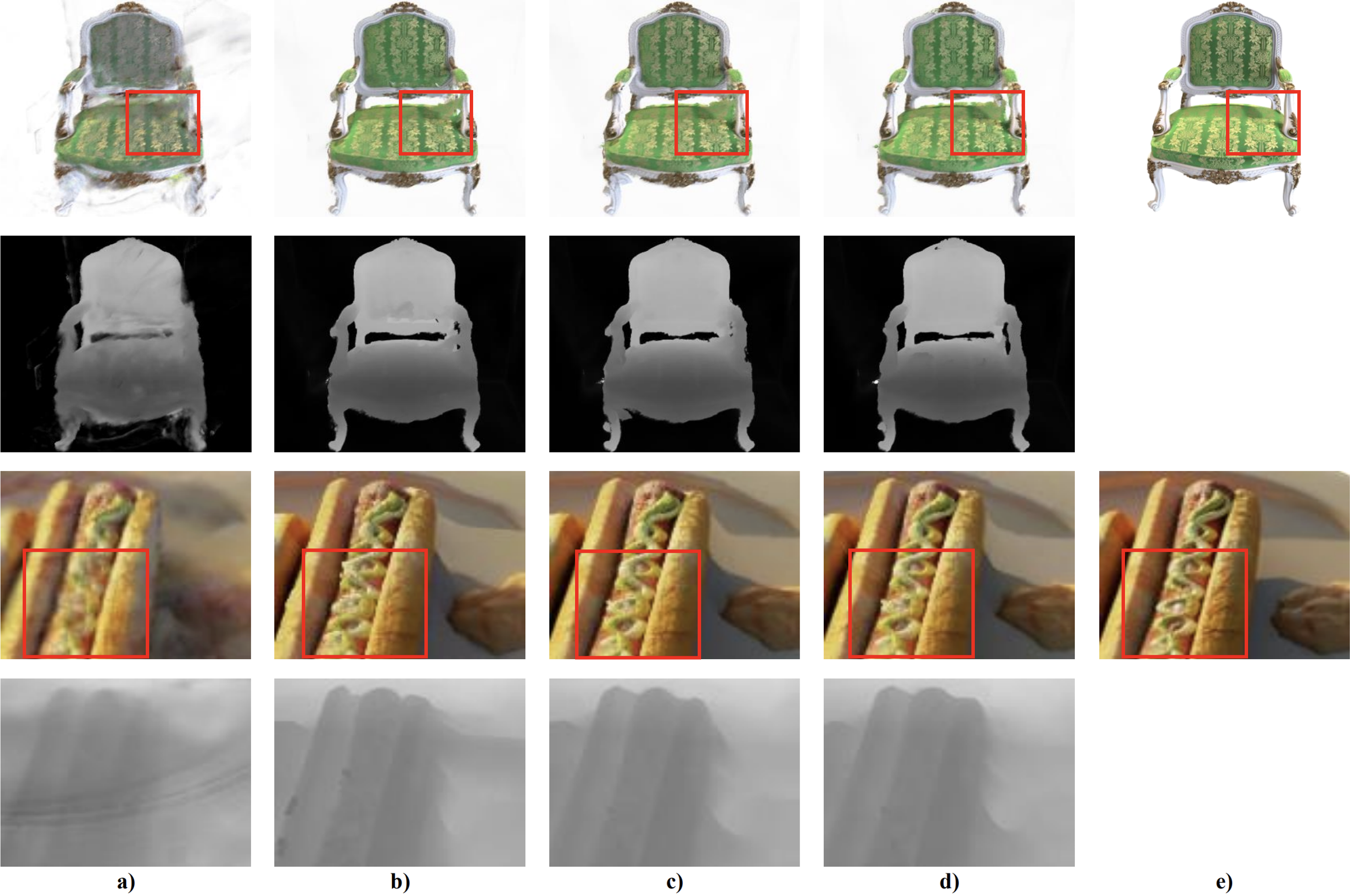}}
\caption{Qualitative results for Vanilla INGP (a), CombiNeRF (b), DWTNeRF with only cross-branch interactions (c), DWTNeRF with full modules (d) and ground-truth (e) on the NeRF-Synthetic \cite{NeRF} dataset (``chair'' and ``hotdog'' scenes)}
\label{fig:vis_NS}
\end{center}
\vskip -0.2in
\end{figure*}

\begin{figure*}[t]
\vskip 0.2in
\begin{center}
\centerline{\includegraphics[width=\textwidth]{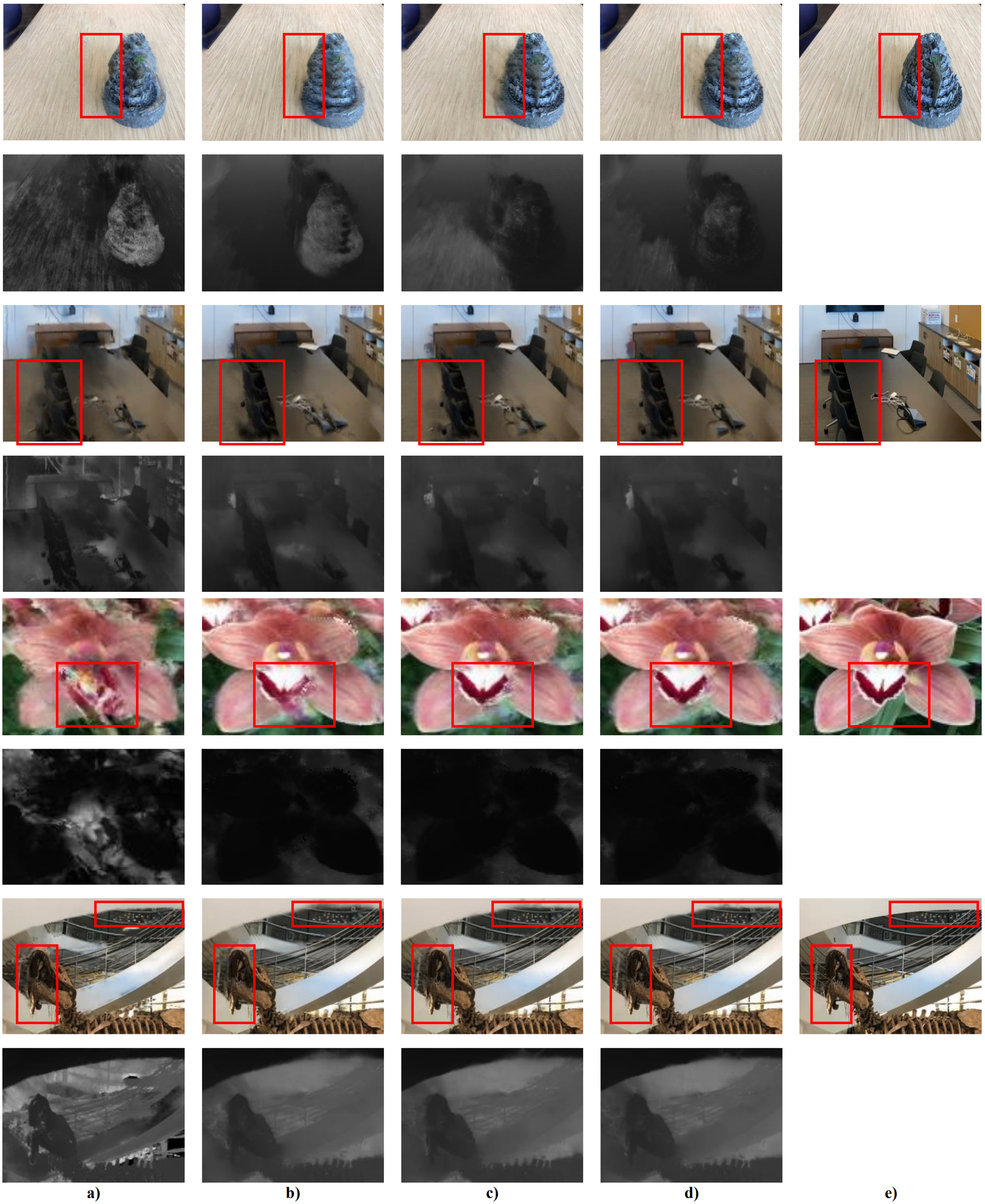}}
\caption{Qualitative results for Vanilla INGP (a), CombiNeRF (b), DWTNeRF with only cross-branch interactions (c), DWTNeRF with full modules (d) and ground-truth (e) on the LLFF \cite{LLFF} dataset. Upper row depicts the novel views, while lower row depicts the corresponding depth
maps. From top to bottom: ``fortress'', ``room'', ``orchids'' and ``trex'' scenes. The  \textbf{\textcolor{Red}{red}} regions show improved visual qualities.} 
\label{fig:vis_LLFF_concat}
\end{center}
\vskip -0.2in
\end{figure*}

\quad We provide more qualitative results on the 3-shot LLFF and 4-shot NeRF-Synthetic benchmarks at Figures \ref{fig:vis_LLFF_concat} and \ref{fig:vis_NS}. There are 6 scenes in total. In each scene, we would like to direct attention towards the regions enclosed by the \textbf{\textcolor{Red}{red}} box, which highlights the effects of cross-branch interactions (b $\rightarrow$ c) and DW loss (c $\rightarrow$ d). They provide finer details, preserve better structures and reduce random ``hallucinated'' points in 3D space.

\section{Wavelet Coefficients} \label{app_sec_wavelet_coeffs}

\begin{table}[h]
\caption{Low-pass filters of the Daubechies \cite{wavelet-db} wavelets}
\label{tab:wavelet_coeffs}
\vskip 0.15in
\begin{center}
\begin{small}
\begin{sc}
\begin{tabular}{lccc}
\toprule
Order & 1 & 2 & 3 \\
\midrule
\multirow{6}{*}{$\ell_k$} 
  & 1           & $1 + \sqrt{3}$   & 0.3327 \\
  & 1           & $3 + \sqrt{3}$   & 0.8069 \\
  &             & $3 - \sqrt{3}$   & 0.4599 \\
  &             & $1 - \sqrt{3}$   & -0.1350 \\
  &             &                  & -0.0854 \\
  &             &                  & 0.0352 \\
\midrule
Factor & $\sqrt{2}$ & $4\sqrt{2}$ & 1 \\
\bottomrule
\end{tabular}
\end{sc}
\end{small}
\end{center}
\vskip -0.1in
\end{table}

Table \ref{tab:wavelet_coeffs} shows the coefficients $\ell_k$ of the 1D low-pass filters $\boldsymbol{\ell}$ that are used in our experiments. The corresponding 2D low-pass matrices $\mathbf{L}$ are constructed with shifting rows of $\boldsymbol{\ell}$:

\begin{equation*} \label{matrix_L}
\mathbf{L} =
\begin{pmatrix}
\cdots & \cdots & \cdots \\
\cdots & \ell_{-1} & \ell_0 & \ell_1 & \cdots \\
& & \cdots & \ell_{-1} & \ell_0 & \ell_1 & \cdots \\
& & & & & \cdots & \cdots \\
\end{pmatrix}
\end{equation*}
and similarly for how $\mathbf{H}$ is constructed with the 1D high-pass filter $\boldsymbol{h}$. We experimented with orthogonal Daubechies wavelets \cite{wavelet-book}, where $\boldsymbol{h}$ is derived trivially from $\boldsymbol{\ell}$ according to the ``Alternating Flip'' condition:

\begin{equation*}
h_k = (-1)^k \ell_{N-1-k}
\end{equation*}
where $k$ denotes a single index of the 1D filter with number of coefficients $N$. As a numerical example, suppose that $\boldsymbol{\ell} = [\ell_0, \ell_1, \ell_2, \ell_3]$, the corresponding $\mathbf{h}$ that meets ``Alternating Flip'' condition is $\boldsymbol{h} = [\ell_3, -\ell_2, \ell_1, -\ell_0]$. In summary, while the DW loss can technically be classified as a prior-based method, we classify it as a regularization-based method as it needs no more priors than the 1D low-pass filter $\boldsymbol{\ell}$. 

\begin{figure*}[ht]
\vskip 0.2in
\begin{center}
\centerline{\includegraphics[width=\textwidth]{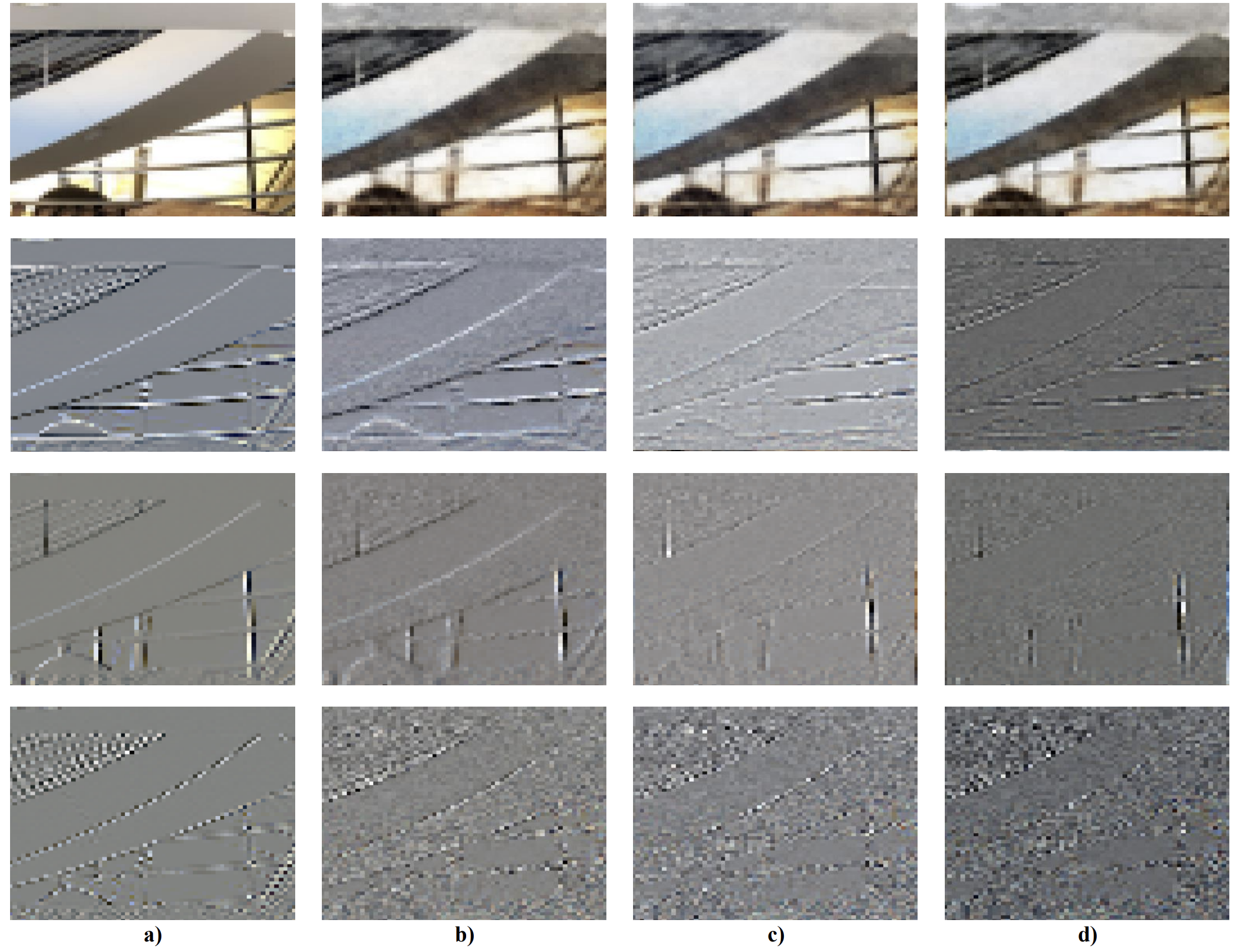}}
\caption{Frequency-based scene decompositions with different wavelets. From top to bottom: the LL, LH, HL and HH sub-bands. The first column (a) shows the sub-bands of ground-truth views using the Haar. The second column (b) is the same, but for rendered views, which are much noisier. The rest show sub-bands of rendered views using the db2 (c) and db3 (d). Higher-order wavelets provide more fine-grained details, but our experiments showed low-frequency supervision with coarse details is sufficient.}
\label{fig:vis_wavelet}
\end{center}
\vskip -0.2in
\end{figure*}

We also provide more visualizations of scene decompositions using different wavelets in Figure \ref{fig:vis_wavelet}. Higher-order wavelets spread energy more evenly across sub-bands, resulting in more fine-grained details. However, as presented in Table \ref{tab:wavelet_abs}, this is not strictly necessary, confirming our intuition that low-frequency supervision is sufficient. While technically the db3 is best-performing, it is costlier due to having more coefficients. Therefore, the Haar was our method of choice.

\section{Model-based methods} \label{app_sec_model}

\quad In this section, we compared DWTNeRF's cross-branch interactions against model-based methods as introduced by mi-MLP \cite{mi-MLP}: ``Residual Connections'' and ``Element-wise Cross-branch Interactions''. The architectures of each method are provided in Figure \ref{arch_mimlp}. In general, model-based methods aim to boost few-shot NeRF by tweaking Vanilla MLP's architecture.

\textbf{Network Structures}. Following CombiNeRF \cite{CombiNeRF} and mi-MLP \cite{mi-MLP}, DWTNeRF is modelled using separate MLP branches for densities and colors. The density branch has 2 MLP layers, and its inputs are encoded using hash encoding \cite{INGP} with 16 resolution levels. The color branch has 3 MLP layers, and its inputs are encoded using 64 bases of the spherical harmonics \cite{Spherical}. 

\textbf{Residual Connections}. This provides a shorter path between the inputs and intermediate MLP layers, helping to reduce overfitting. For our implementation onto INGP, we introduced the residuals with the concatenation operator $[\cdot,\cdot]$:

\begin{equation*}
    f_i = \texttt{Linear}([f_{i-1},\gamma(\cdot)])
\end{equation*}
where \texttt{Linear} is Pytorch's Linear layer. $f_i$ and $f_{i-1}$ denote the outputs of the $i$-th layer and the layer preceding it. The encoded inputs $\gamma(\cdot)$ are either $\gamma(\mathbf{x})$ or $\gamma(\mathbf{d})$, depending on the MLP branch. We implemented residuals for both branches, up until the penultimate layer. 

\textbf{Element-wise Cross-branch Interactions}. This provides interactions between color and density branches, regularizing color predictions through position-aware relationships. We implemented this into INGP as follows:

\begin{equation*}
f^{\mathbf{d}}_{i}=\texttt{Linear}_1[f^{\mathbf{d}}_{i-1}+\texttt{Linear}_2(f^{\mathbf{x}}_{i-1})]
\end{equation*}
where $f^{\mathbf{d}}_{i}$ and $f^{\mathbf{x}}_{i}$ denote the $i$-th layer at the color and density branches, respectively. The second linear layer, $\texttt{Linear}_2$, is to ensure similar dimensions between the intermediate features from each branch for element-wise addition. The interaction is one-way (i.e., from the density towards the color branch, but not vice versa) to keep the density features lean. This is because densities are only position-dependent, and are vulnerable to over-parameterization.

As presented in Table \ref{tab:prior_and_model_abs}, all of the above methods degrade performance. Similar to some prior-based methods, model-based methods are not trivially transferable across Vanilla NeRF and INGP, and very possibly, 3DGS. This is due to INGP's aggressive optimization dynamics, making it vulnerable to very slight architectural changes. Our method, based on multi-head attention and acting only on the inputs \& outputs, bypasses this shortcoming while still capable of learning density-color interactions.

\end{document}